\documentclass{isprs} 
\usepackage{subfigure}
\usepackage{setspace}
\usepackage{geometry} 
\usepackage{epstopdf}
\usepackage[labelsep=period]{caption}  
\usepackage[british]{babel} 
\usepackage[hang]{footmisc}
\usepackage{tabularray}
\usepackage{amsmath}

\begin{document}

\title{SCP: SCENE COMPLETION PRE-TRAINING FOR 3D OBJECT DETECTION}
\date{}


 
\author{ 
 Yiming Shan\textsuperscript{1\footnotemark[2]}, Yan Xia\textsuperscript{1, 2 \footnotemark[2]} \thanks{Corresponding author} ,  Yuhong Chen\textsuperscript{1}, Daniel Cremers\textsuperscript{1,2}}

\address{
	\textsuperscript{1 }Technical University of Munich, Germany\\
 	\textsuperscript{2 }Munich Center for Machine Learning (MCML), Germany\\
  (yiming.shan, yan.xia, yuhong.chen, cremers)@tum.de
}


\commission{XX, }{YY} 
\workinggroup{XX/YY} 
\icwg{}   

\abstract{
3D object detection using LiDAR point clouds is a fundamental task in the fields of computer vision, robotics, and autonomous driving. However, existing 3D detectors heavily rely on annotated datasets, which are both time-consuming and prone to errors during the process of labeling 3D bounding boxes. In this paper, we propose a Scene Completion Pre-training (SCP) method to enhance the performance of 3D object detectors with less labeled data. SCP offers three key advantages: (1) Improved initialization of the point cloud model. By completing the scene point clouds, SCP effectively captures the spatial and semantic relationships among objects within urban environments. (2) Elimination of the need for additional datasets. SCP serves as a valuable auxiliary network that does not impose any additional efforts or data requirements on the 3D detectors. (3) Reduction of the amount of labeled data for detection. With the help of SCP, the existing state-of-the-art 3D detectors can achieve comparable performance while only relying on 20\% labeled data. 
}

\keywords{LiDAR Point Clouds, 3D Object Detection, Pre-training, Scene Completion, Autonomous Driving}

\maketitle
\renewcommand{\thefootnote}{\fnsymbol{footnote}}
\footnotetext[2]{Equal contribution}

\section{Introduction}\label{MANUSCRIPT}
 
\sloppy

3D object detection using LiDAR point clouds is a key task in the domains of computer vision~\cite{mao2021voxel}, robotics~\cite{zhou2018voxelnet}, and autonomous driving~\cite{yan2018second}. In contrast to 2D images, point clouds obtained through mobile laser scanning (MLS) offer accurate 3D geometric properties and depth insights~\cite{xia2021vpc}, endowing them with superior resilience for object detection under diverse illumination conditions~\cite{xia2023lightweight}. Over the past few years, numerous learning-based 3D detection techniques have exhibited remarkable performance by leveraging extensive supervised training on large annotated datasets. Nevertheless, the annotation of point clouds presents a substantial challenge due to (1) inherent incompleteness and occlusion, which render the identification of points ambiguous~\cite{li2023fast}; and (2) the time-consuming and error-prone nature of labeling individual points or delineating 3D bounding boxes~\cite{wang2021unsupervised}.

A possible solution is to learn 3D model initialization using an unsupervised pre-training way and then fine-tune the models with small labeled data. A recent line of pre-training works based on generative adversarial networks (GANs) is proposed. ~\cite{sauder2019self} proposes to learn the rearrangement of the point clouds by predicting the original voxel location of each point. However, this approach is incapable of handling rotated and translated point clouds effectively due to the permutation variability exhibited in their voxel representations. Furthermore, \cite{sharma2020self} explores to classify each point into the assigned partitions based on cover trees. However, it ignores the semantically contiguous regions(e.g., airplane wings, car tires). In addition, PointContrast~\cite{xie2020pointcontrast}  leverages established point-wise correspondences between various views of a complete 3D scene to pre-train weights for point clouds. However, this approach may not be suitable for dynamic urban environments. Recently, a novel approach has been proposed to learn model initialization by completing the 3D shape of single objects~\cite{wang2021unsupervised}. This method demonstrates notable advancements in various downstream tasks, such as object classification and segmentation, by effectively completing individual objects. However, it overlooks the significance of both spatial and semantic relationships among objects, which are crucial considerations for successful 3D detection tasks in complex urban environments.

To tackle this problem, we propose a novel Scene Completion Pre-training network, named SCP, aiming to learn a robust model initialization for 3D object detection from single LiDAR scans. Our SCP involves training a voxel-based scene completion network consisting of a feature encoder and a decoder. The encoder utilizes a Transformer-based 3D backbone~\cite{mao2021voxel} to efficiently extract informative features from the raw point clouds. Concurrently, the decoder incorporates an anisotropic convolution (AIC) module~\cite{li2020anisotropic}, which dynamically adapts the receptive fields for different voxels. 
By completing the scene point cloud, our method accurately learns the spatial and semantic relationships among the objects. This enables the pre-training model to serve as an effective model initialization when employed as the 3D backbone in a 3D detection network.  Furthermore, a small labeled data is introduced to fine-tune the 3D detection network.

To summarize, the main contributions of this work are:
\begin{itemize}
    \item We propose a voxel-based Scene Completion Pre-training network, called SCP, which purely applies to LiDAR point clouds. With the help of the carefully designed encoder and decoder, SCP provides a robust model initialization for the next detection network, encoding the spatial and semantic relationships within urban environments.
    \item We conduct extensive experiments on the KITTI 3D detection benchmark~\cite{geiger2013vision} to demonstrate the effectiveness of our SCP. Notably, the existing state-of-the-art methods with SCP yield comparable performance while relying on 20\% labeled data.
\end{itemize}






\begin{figure*}[ht!]
\begin{center}
		\includegraphics[width=2\columnwidth]{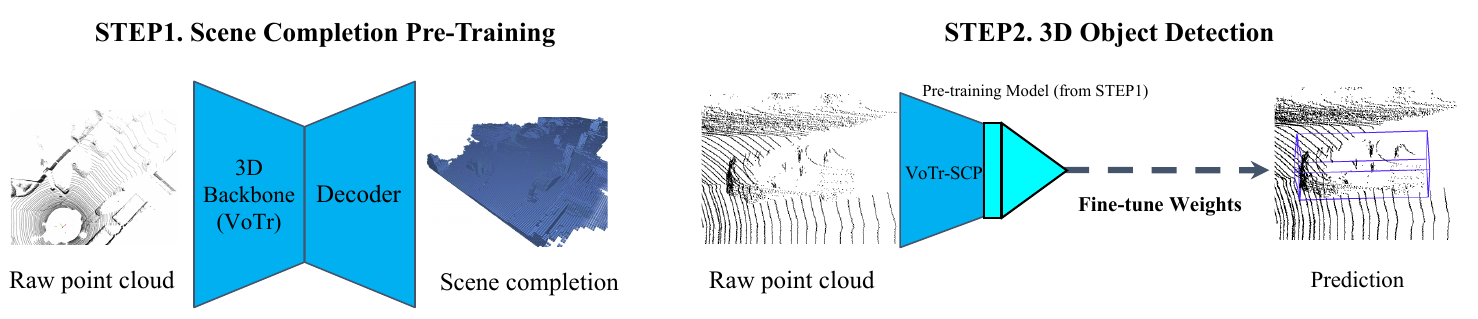}
	\caption{Overview of our SCP. 
Step 1: An encoder-decoder scene completion model is designed to complete the raw point cloud. 
Step 2: Pre-training model weights from step 1 are used for the following 3D detection task.
}
\label{fig:overview}
\end{center}
\end{figure*}

\section{Related Work}\label{sec: related work}
In this section, we give a brief literature review of 3D object detection and scene completion, respectively. 

\textbf{3D object detection.} Early 3D object detection works~\cite{shi2019pointrcnn,yang20203dssd,mao2021voxel,zhou2018voxelnet,yan2018second,shi2020pv,xia2023lightweight} can be broadly categorized into two main methods: the point-based and the voxel-based detectors. The point-based approaches focus on directly capturing features and predicting 3D bounding boxes from the raw points. 
PointRCNN~\cite{shi2019pointrcnn} extracts features from the foreground points and derives the corresponding 3D bounding box. 3DSSD~\cite{yang20203dssd} removes the FP layer and refinement module to reduce computational complexity and proposes a new fusion sampling strategy that yields improved results using fewer representative points. 
VoTr~\cite{mao2021voxel} introduces a voxel transformer-based 3D detection backbone, presenting an alternative solution to the task of 3D object detection. 
The voxel-based approach in 3D object detection involves transforming the large and non-structured point cloud data into voxels, which enables efficient feature extraction and saves computational time. 
VoxelNet~\cite{zhou2018voxelnet} and SECOND~\cite{yan2018second}, for instance, partition the points into voxels and utilize 3D sparse convolution to extract features. 
Subsequently, they employ the Region Proposal Network (RPN) to obtain 3D bounding boxes. 
On the other hand, PV-RCNN~\cite{shi2020pv} combines the strengths of both approaches. 
It leverages multi-scale techniques to generate high-quality proposals from voxel-based methods while also incorporating fine-grained local information from point-based methods. 
Recently, DMT~\cite{xia2023lightweight} explores motion prior knowledge to generate accurate 3D positions and rotation.


\textbf{3D scene completion.}  Early works on 3D completion mainly focus on single objects~\cite{xia2021asfm,wang2022learning,xia2021vpc}. Comparably, completing the whole scene poses greater challenges since the scene point cloud is large-scale and has many objects with various densities. The pioneering work by Song\cite{song2017semantic} explores the depth maps for 3D scene completion and leverages the scene information derived from the depth map for semantic segmentation. Scene completion and semantic segmentation are closely intertwined tasks, and jointly processing them can yield mutual performance improvements. JS3C~\cite{yan2021sparse} introduces the Point-Voxel Interaction (PVI) module to enhance knowledge fusion between the semantic segmentation and semantic scene completion tasks. This module facilitates interaction between incomplete local geometries in point clouds and complete global structures in voxels, enabling a more comprehensive understanding of the scene. AICNet~\cite{li2020anisotropic} proposes a novel anisotropic convolution, which decomposes a 3D convolution into three consecutive 1D convolutions.  S3CNet~\cite{cheng2021s3cnet} tackles the challenge of large-scale environments by incorporating sparsity considerations and leveraging a sparse convolution-based neural network. Recently, SCPNet~\cite{xia2023scpnet} introduces a novel knowledge distillation objective termed as Dense-to-Sparse Knowledge Distillation (DSKD).

\section{Method}\label{MANUSCRIPT}
The overview of our SCP for 3D object detection can be divided into two stages, as illustrated in Fig.~\ref{fig:overview}. 
In the first step (scene completion pre-training), we employ an encode-decode model to effectively complete the partial scene point clouds. This involves leveraging available data to predict and generate the missing parts of the point cloud, resulting in a more comprehensive representation of the scene.
In the second step (3D object detection), we utilize the learned weights from the scene completion pre-training model as an initialization for the 3D detectors. By transferring the knowledge acquired during the scene completion, we establish a strong spatial and semantic relationship of objects, leading to improved detection performance and efficiency, especially in the case of smaller labeled data.

\begin{figure}[ht!]

\begin{center}
\includegraphics[width=0.99\columnwidth]{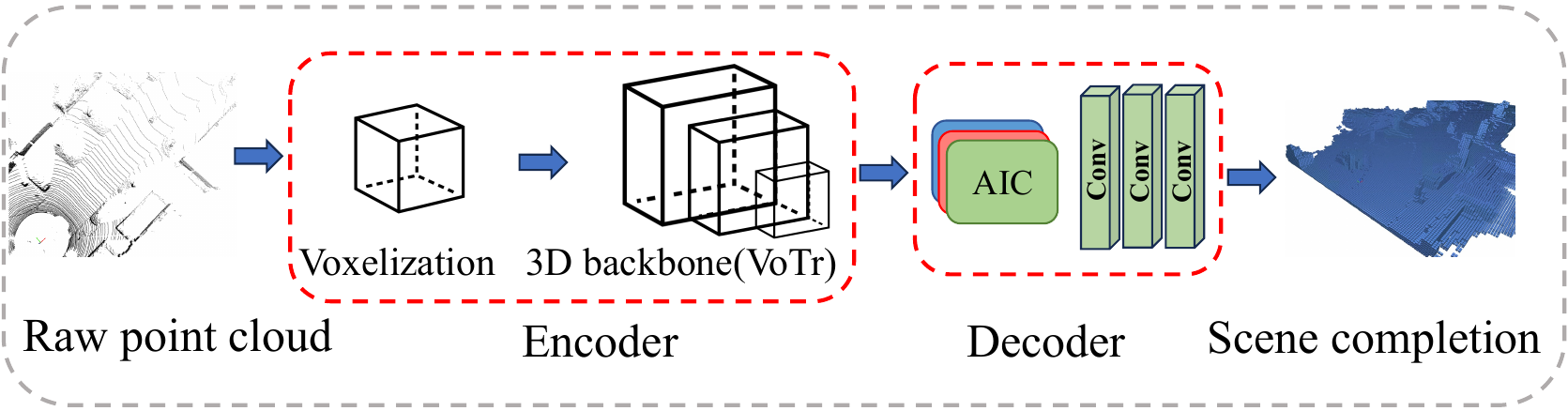}
\caption{The pipeline of scene completion network.
The encoder is a Transformer-based backbone and the decoder includes an AIC module and three convolutional layers.}
\label{fig:scp_pipeline}

\end{center}
\end{figure}

\paragraph{Scene completion pre-training}
Next, we provide a detailed pipeline of SCP in Fig. \ref{fig:scp_pipeline}. 
This section is divided into four stages: voxelization, encoder, decoder, and prediction.

\textbf{a) Voxelization.} The raw point cloud is initially transformed into structured voxels, which serve to facilitate the subsequent feature extraction process.

\begin{figure}[ht!]
\begin{center}
	\includegraphics[width=0.65\columnwidth]{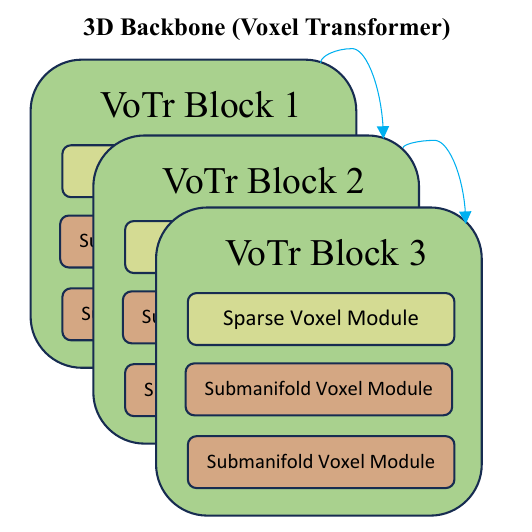}
	\caption{The architecture of the 3D backbone network. It consists of three VoTr blocks, each layer containing two different self-attention modules.} 
\label{fig:votr}
\end{center}
\end{figure}

\textbf{b) Encoder.} 
We use a transformer-based 3D backbone network to extract features from voxels.
The architecture of the 3D backbone is the same as VoTr~\cite{mao2021voxel}, as illustrated in Fig.~\ref{fig:votr}.
The voxel undergoes a sequence of three "VoTr Block" layers. Each block layer consists of one sparse voxel module and two submanifold voxel modules. As the voxel passes through each block layer, its features are effectively extracted, and the voxel is downsampled three times.
In these voxels, both non-empty voxels and empty voxels are present. The submanifold voxel modules are designed to handle the non-empty voxels and utilize self-attention mechanisms~\cite{xia2021soe,Xia_2023_ICCV} to effectively extract features from them. On the other hand, the sparse voxel modules are specifically designed for empty voxels, allowing them to perform feature extraction on these regions.

\textbf{c) Decoder.}
Subsequently, the features obtained from the encoder are fed into the decoder, which includes Anisotropic Convolutional Networks~\cite{li2020anisotropic} and three layers of 3D convolution. 
Instead of employing traditional 3D convolution directly, the AIC model decomposes the 3D convolution into three separate operations, each corresponding to one dimension ($x$, $y$, $z$). Within each dimension, three 1D convolutions are inserted based on modulation factors, represented by a $3\times3$ matrix. The modulation factors matrix contains information about nine different weights in X, Y, and Z three different dimensions (K\_x1, K\_x2... K\_z3).
These modulation factors are all positive, and the sum of each row(dimension) equals one.
By utilizing distinct convolution kernels and sizes for each 1D convolution, the AIC model enables improved incorporation of geometric information and provides enhanced flexibility. Ultimately enhancing the performance of the model.


\textbf{d) Prediction.} The completion predictions obtained from this pipeline represent a comprehensive and coherent depiction of the entire scene, effectively filling in the gaps and providing a more holistic representation of the raw data.

\begin{figure}[t]
\begin{center}
		\includegraphics[width=0.99\columnwidth]{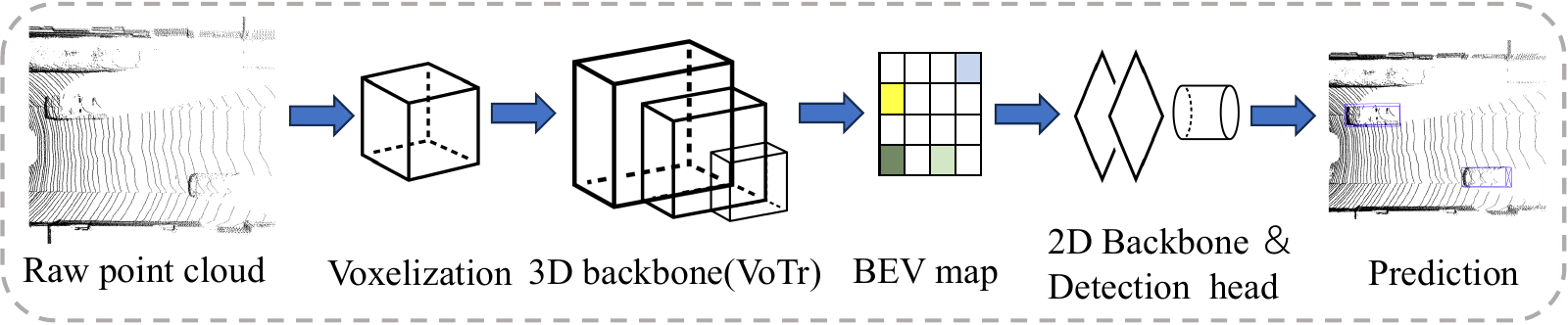}
	\caption{The pipeline of the 3D detector VoTr. VoTr mainly contains voxelization, 3D backbone, BEV map, 2D backbone, and detection head modules.}
\label{fig:votr_pipeline}
\end{center}
\end{figure}

\paragraph{3D object detection.}
In the first step, we acquired well-performing pre-training weights, which are then utilized to aid in 3D detector~\cite{mao2021voxel}. The pipeline of the detector VoTr is shown in Fig.~\ref{fig:votr_pipeline}.
This knowledge transfer enables the utilization of valuable insights gained from scene completion to improve the accuracy and reliability of 3D detection results.
In the initial step, the raw point cloud data undergoes voxelization, converting them into the structured voxel representation. These voxelized data are then fed into the 3D backbone, which is the same in the scene completion.
The extracted features are then projected onto a bird's-eye view (BEV) map to generate 3D proposals, followed by the utilization of a 2D backbone and a detection head for further processing.
By leveraging the learned knowledge from scene completion, the pre-training weights gain a deeper understanding of the scene, resulting in improved accuracy and reliability in the 3D detection results.

\section{Experiments}\label{MANUSCRIPT}
This section first provides a comprehensive description of the datasets used for both the pre-training and fine-tuning detection stages. Subsequently, we present and discuss the detection results, shedding light on the performance and effectiveness of our SCP.

\subsection{Datasets}\label{sec:Abstract}

{\bf SemanticKITTI dataset.} We utilize SemanticKITTI~\cite{behley2019iccv} as the pre-training dataset. SemanticKITTI is a pioneering real-world outdoor benchmark specifically designed for 3D semantic scene completion. The dataset consists of volumetric representations of the 3D scenes, with dimensions of $256\times256\times32$ voxels. Empty voxels, which do not contain any points, are labeled accordingly. For our training phase, we utilize a total of 19,130 pairs of input and target voxel grids, while reserving 815 pairs for validation purposes.


{\bf KITTI detection benchmark.} 
To evaluate the effectiveness of our SCP in 3D object detection, we use the KITTI benchmark~\cite{geiger2013vision}, which comprises 7,481 samples for training and 7,518 for testing. Following VoTr~\cite{mao2021voxel}, the training dataset is further subdivided into 3,712 samples for training data and 3,769 samples for validation data~\cite{chen20153d}. 



\subsection{Evaluation metric }\label{sec:Abstract}

Following \cite{hossin2015review}, we first employ the Intersection over Union (IoU) metric for the 3D scene completion. In 3D object detection, we then use the Average Precision (AP) with 11 recall points (AP11) for performance assessment.


{\bf Intersection over union (IoU).} 
IoU is a metric used to measure the overlap between a model's predictions and the ground truth. It quantifies the degree of alignment between the two sets. An IoU of 0 indicates no overlap, meaning there is no intersection between the predicted regions and the ground truth. Conversely, an IoU of 1 represents a complete overlap, where the predicted regions perfectly match the ground truth. In practice, a larger IoU indicates better algorithm performance. The IoU is calculated as follows:

\begin{equation}
        IoU=\frac{TP}{TP+FP+FN},
\end{equation}
\noindent
where TP (True Positive) denotes the correct prediction of the ground truth. FP (False Positive) denotes incorrectly predicting an object that is not in the ground truth, and FN (False Negative) denotes the ground truth not being predicted.

{\bf Average precision 11(AP11).} 
The Average Precision (AP) is computed by integrating the Precision-Recall (PR) curve. To approximate the PR curve, the 11-point interpolation method is commonly employed.
In the 11-point interpolation, the Precision-Recall curve is sampled at 11 equally spaced recall levels between 0 and 1. The formula for AP11 is as follows:
\begin{equation}
AP_{11}=\frac{1}{11}\sum_{R\in\{0,0.1, \ldots, 0.9,1\}}P_{\text {interp}}(R)
\end{equation}
where $P_{\text {interp}}(R)$ is the maximum precision for recall greater than R (Recall).

\subsection{Scene completion pre-training }\label{sec:Abstract}
\begin{figure}[htbp]
\begin{center}
		\includegraphics[width=0.99\columnwidth]{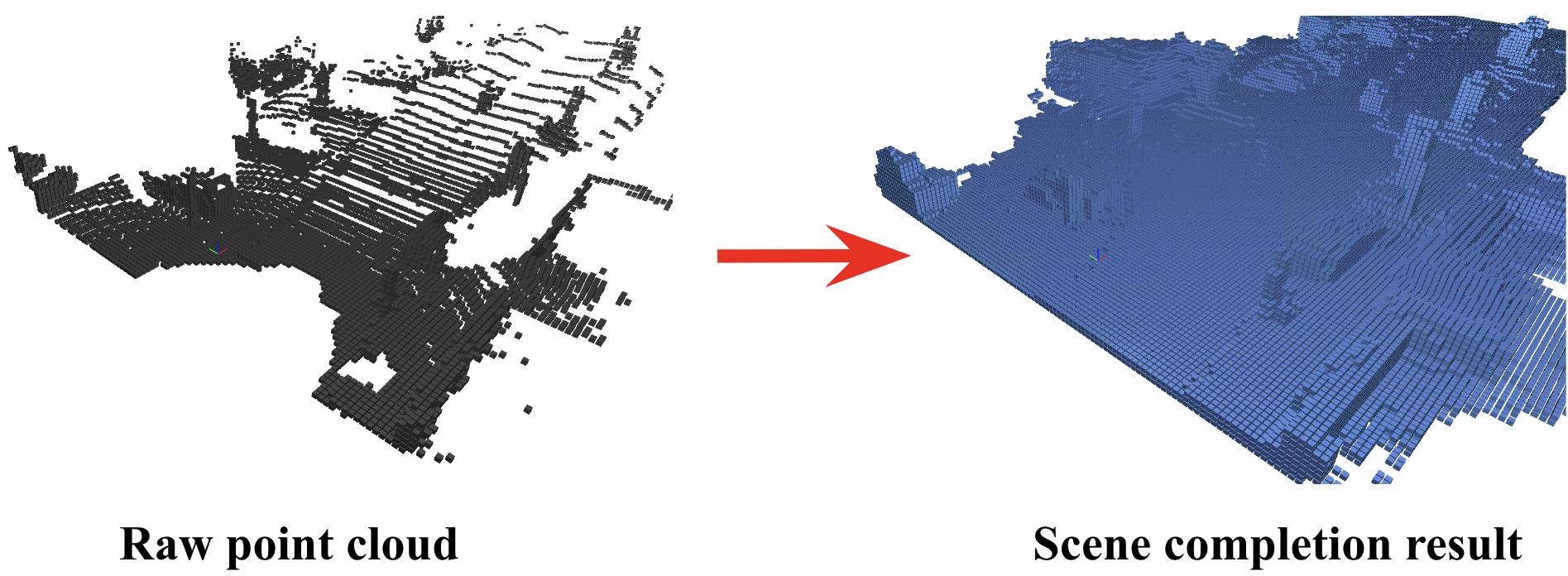}
	\caption{The comparison before and after the scene completion.}
\label{fig:sc}
\end{center}
\end{figure}

The raw point cloud serves as input and is fed into our SCP. Notably, only 20\% of the training data is sampled from the overall training set for training the SCP. Through this step, we obtain the pre-training model for scene completion.
As depicted in Fig.~\ref{fig:sc}, it is evident that after pre-training the complementary network, the original data is effectively completed and generates a more holistic representation of the scene.

\subsection{3D object detection}\label{sec:Abstract}
During this stage, the objective is to train and optimize the components of the detection network that complement the fully trained 3D backbone.
The pre-training weights in the scene completion provide a solid starting point and enable efficient transfer of prior knowledge. By building upon the already well-performing 3D backbone part, we can concentrate on refining and fine-tuning the other components of the detection network. 
To evaluate the effectiveness of our SCP, we employ the entire validation dataset and compare it with the fully trained 3D detectors.

\subsection{Implementation details}\label{sec:Abstract}
The pre-training scene completion module employed the Adam optimizer with a batch size of 3 and an initial learning rate of 0.001. The learning rate dynamically changed during training based on the number of training rounds. The training process utilized 20\% of the SemanticKITTI dataset, with approximately 50 epochs of training conducted.
Regarding the 3D detection module, the Adam-one cycle optimizer was employed with a batch size of 6 for SCP-SSD and 3 for SCP-TSD. The learning rate was set to 0.003. The SCP-SSD model was trained for 100 epochs, while the SCP-TSD model was trained for 80 epochs.
All experiments and training processes were conducted on GPUs, specifically using hardware such as the Quadro RTX 8000 and NVIDIA A40. These GPUs offer high-performance computing capabilities and possess a substantial memory capacity of 48GB.

\subsection{Results}\label{sec:Abstract}

We conduct a comparative analysis by assembling our SCP and the state-of-the-art 3D detector VoTr~\cite{mao2021voxel}. To ensure a fair comparison, we adopt the same evaluation metrics, following~\cite{mao2021voxel}. 
Table.~\ref{tab:Margin_settings} presents the AP11 achieved by each method on the KITTI validation split for the car category. 
Table.~\ref{tab:iou_sc} illustrates the results of scene completion.
Despite the utilization of a relatively small amount (20\%) of training data, our SCP exhibits remarkable performance in object detection tasks. It is worth highlighting that our method achieves detection results that are remarkably close to those obtained through full data training, even surpassing the performance of full-data training.


\paragraph{3D scene completion.}
In this study, we utilized only 20\% of the SemanticKITTI dataset for training the completion network. 
The completion results obtained in our experiments ranged from 47.39\% to 54.58\%. These results reflect the effectiveness of our completion network in generating accurate and reliable scene completions.
Notably, our pre-training model, which underwent rigorous training and optimization, yielded the highest completion result of \textbf{54.58\%}.
The training results of the scene completion network are shown in Fig.~\ref{tab:iou_sc}.

\begin{table}[ht!]
\centering
\caption{The IoU results of 3D scene completion in different training epochs.}
\label{tab:iou_sc}
\begin{tblr}{
  cells = {c},
  vlines,
  hline{1,2,3,4,5,6,7,8} = {-}{},
  hline{2} = {2-4}{},
}
Training rounds              & IoU(\%)                              \\
    
epochs=10            & 51.67                  \\
epochs=20            & 52.11         \\
epochs=30            & 53.79                 \\
epochs=40            & 54.14  \\
epochs=50            & 54.48  
\end{tblr}
\end{table}

\paragraph{3D object detection.}

In this experiment, we focus on training the VoTr-SCP model using a significantly reduced amount of training data. Specifically, we utilized only 20\% of the available data for training purposes. This deliberate reduction in training data allowed us to investigate the model's performance under resource-constrained scenarios and assess its ability to leverage limited data effectively. The results are presented in Table.~\ref{tab:Margin_settings}.
Remarkably, the reduction in training data does not lead to a significant degradation in the detection performance of the VoTr-SCP model. Despite utilizing a smaller amount of training data, our approach maintains a highly effective detection performance that is comparable to, and in some cases even surpasses, the results obtained through full data training. Here, we have only calculated the detection results on the car category.

\begin{figure}[ht!]
\begin{center}
		\includegraphics[width=0.8\columnwidth]{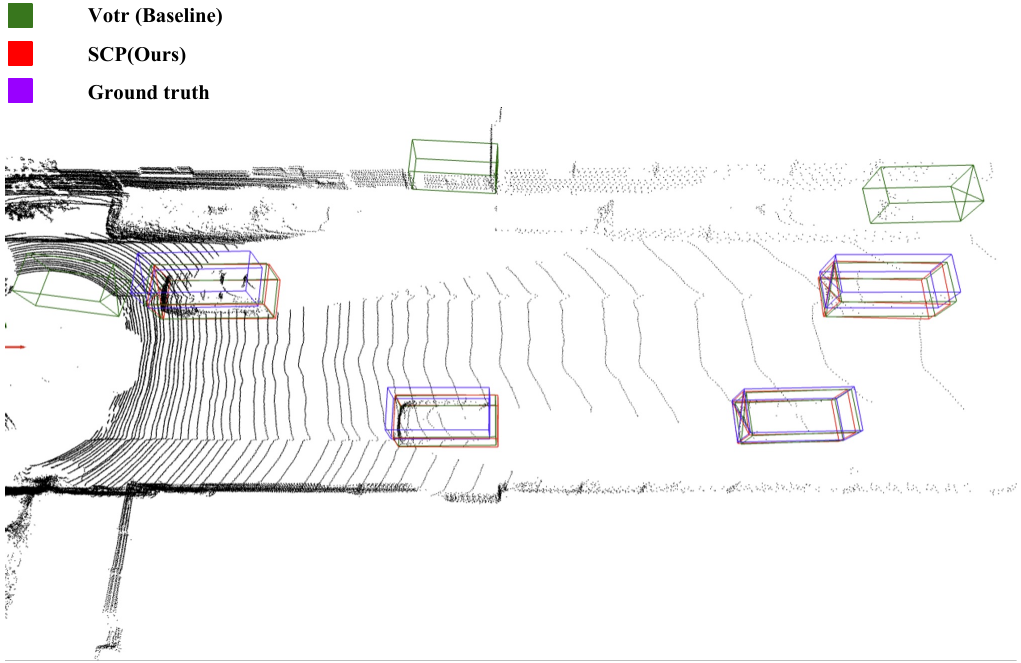}
	\caption{Visualization of the example results. Due to the existence of some points similar to cars in the target area, VoTr incorrectly identified the wall as a car (the green bounding box), whereas our SCP can help to reduce these errors.
}
\label{fig:detection}
\end{center}
\end{figure}

Specifically, our SCP demonstrates notable improvements in the easy car class, elevating the performance from 89.04 to \textbf{89.10} compared to VoTr-TSD. This finding highlights the effectiveness of our approach in enhancing the detection results. Additionally, we provide a visualization of a scene completion result in Fig.~\ref{fig:detection}.

\begin{table}[ht!]
\centering
\caption{Comparisons on the KITTI with AP11 for the car category (20\% training data).}
\label{tab:Margin_settings}
\begin{tblr}{
  cells = {c},
  cell{1}{1} = {r=2}{},
  cell{1}{2} = {c=3}{},
  vlines,
  hline{1,3,5,7} = {-}{},
  hline{2} = {2-4}{},
}
Methods             & AP11(\%)       &       &                \\
                    & Easy           & Mod   & Hard           \\
VoTr-SSD~\cite{mao2021voxel}            & 87.86          & 78.27 & 76.93          \\
VoTr-SSD-SCP (Ours) & {86.86}      &{75.75}   &{68.79}     \\
VoTr-TSD~\cite{mao2021voxel}            & 89.04          & 84.04 & 78.68          \\
VoTr-TSD-SCP (Ours) & \textbf{89.10} &{78.78} &{78.04} 
\end{tblr}
\end{table}

\section{Discussions}\label{MANUSCRIPT}
\paragraph{Different labeled data volumes.} 
To comprehensively assess the influence of the pre-training model on the detection performance under different amounts of labeled data, we carried out experiments utilizing varying fractions of the training set. Specifically, we conducted experiments using four different fractions: 10\%, 20\%, 30\%, and 40\% of the total training set.

The results of these experiments are presented in Table.~\ref{tab:ssd_data_volume} and Table.~\ref{tab:tsd_data_volume}. It is evident from the tables that as the volume of data increases, the performance of the models improves.
Notably, our experimental results highlight the robustness and efficacy of our model, even when trained with very small amounts of data. At data volumes as low as 10\% and 20\% of the training set, our model consistently achieved commendable results. 

\begin{table}[ht!]
\centering
\caption{Comparisons on different data volume with VoTr-SSD-SCP (IoU=51.6) for the car category.}
\label{tab:ssd_data_volume}
\begin{tblr}{
  cells = {c},
  cell{1}{1} = {r=2}{},
  cell{1}{2} = {c=3}{},
  vlines,
  hline{1,3,4,5,6,7} = {-}{},
  hline{2} = {2-4}{},
}
Data Volume             & AP11(\%)       &       &                \\
                    & Easy           & Mod   & Hard           \\
10\%            & 85.34 & 71.89 & 67.85\\
20\%            & 85.91 & 72.36& 68.25          \\
30\%            & 86.43 & 74.43& 68.28           \\
40\%            & \textbf{86.79} & \textbf{76.02} & \textbf{72.01}          \\
\end{tblr}
\end{table}

\begin{table}[ht!]
\centering
\caption{Comparisons on different data volume with VoTr-TSD-SCP (IoU=54.1) for the car category.}
\label{tab:tsd_data_volume}
\begin{tblr}{
  cells = {c},
  cell{1}{1} = {r=2}{},
  cell{1}{2} = {c=3}{},
  vlines,
  hline{1,3,4,5,6,7} = {-}{},
  hline{2} = {2-4}{},
}
Data Volume              & AP11(\%)       &       &                \\
                    & Easy           & Mod   & Hard           \\
10\%            & 88.11 & 77.90 & 76.84       \\
20\%            & \textbf{89.10} & 78.78 & 78.04    \\
30\%            & 88.50 & 78.43 & 77.67          \\
40\%            & 89.06 & \textbf{78.95} & \textbf{78.44}
\end{tblr}
\end{table}

\paragraph{Different completion effects.}
To investigate the influence of scene completions on the 3D detection performance, we conducted an in-depth analysis by selecting two different completion results with varying intersection over union (IoU). Specifically, we chose one completion result with an IoU of 51.6\% and another with an IoU of 54.1\%. We then compared the performance of these completion results across different volume datasets.
Table.\ref{tab:iou_ssd} and Table.\ref{tab:iou_tsd} demonstrate that different scene completion results yield varying effects on 3D detection for the VoTr-SSD-SCP and VoTr-TSD-SCP frameworks, respectively. Notably, when utilizing the completion results with better performance (IoU=54.1\%), the detection results show significant improvements on the 10\% and 20\% datasets.
We thus get two conclusions: (1) Accurate and reliable scene completions play a critical role in improving the overall detection performance, particularly in scenarios with very limited data. (2) As the completion achieves better results, it becomes increasingly beneficial for 3D object detection.

\begin{table}[ht!]
\centering
\caption{Comparisons on different IoU with VoTr-SSD-SCP for the car category.}
\label{tab:iou_ssd}

\begin{tblr}{
  cells = {c},
  cell{1}{1} = {r=2}{},
  cell{1}{2} = {c=3}{},
  vlines,
  hline{1,3,5,7,9,11} = {-}{},
  hline{2} = {2-4}{},
}
Completion Results             & AP11(\%)       &       &                \\
                    & Easy           & Mod          & Hard           \\
IoU=51.6(10\%)            & 85.34 & 71.89 & \textbf{67.85}\\
IoU=54.1(10\%)            & \textbf{85.73}        & \textbf{72.63}        & 67.38           \\
IoU=51.6(20\%)            & 85.91& 72.36& 68.25          \\
IoU=54.1(20\%)            & \textbf{86.86}      &\textbf{75.75}   &\textbf{68.79}         

\end{tblr}
\end{table}


\begin{table}[h]
\centering
\caption{Comparisons on different IoU with VoTr-TSD-SCP for the car category.}
\label{tab:iou_tsd}

\begin{tblr}{
  cells = {c},
  cell{1}{1} = {r=2}{},
  cell{1}{2} = {c=3}{},
  vlines,
  hline{1,3,5,7,9,11} = {-}{},
  hline{2} = {2-4}{},
}
Completion Results             & AP11(\%)       &       &                \\
                    & Easy           & Mod          & Hard           \\
IoU=51.6(10\%)           & \textbf{88.23} & \textbf{77.95} & 76.81          \\
IoU=54.1(10\%)            & 88.11 &77.90 &\textbf{76.84}                    \\
IoU=51.6(20\%)            & 88.43& 78.43& 77.57         \\
IoU=54.1(20\%)            & \textbf{89.10} &\textbf{78.78} &\textbf{78.04}          
\end{tblr}
\end{table}



\section{Conclusions}\label{MANUSCRIPT}
In this paper, we propose SCP, a novel scene completion pre-training network for 3D object detection. We have demonstrated that scene completion can learn a model initialization to help the 3D detectors only trained on a small amount of dataset. Experiments indicate that the quality of the scene completion has a positive correlation with the effectiveness of object detection. In the future, we hope to explore the scene completion pre-training into more downstream tasks, for example, 3D semantic segmentations and 3D forecasting. 

{
	\begin{spacing}{1.17}
		\normalsize
		\bibliography{ISPRSguidelines_authors} 
	\end{spacing}
}

\end{document}